%% file: arxiv.tex
\documentclass{article} 
\usepackage{iclr2027_conference,times}

\input{math_commands.tex}

\usepackage{graphicx} 
\usepackage{caption}  
\usepackage{tabularx}
\usepackage{adjustbox}
\usepackage{subcaption}
\usepackage{amsmath}
\usepackage{hyperref}
\usepackage{comment}
\usepackage{cleveref}
\usepackage{xspace}
\usepackage{float}
\usepackage{bbding}
\usepackage{multirow}
\usepackage{xcolor} 
\usepackage{pifont}
 
\usepackage{amssymb}
\usepackage{soul}
\usepackage{graphicx}
\usepackage{booktabs}
\usepackage{tcolorbox}
\usepackage{booktabs}
\usepackage{enumitem}
\usepackage{wrapfig}
\tcbuselibrary{breakable}
\definecolor{stmmdoorbell}{HTML}{1F77B4}
\definecolor{stmmalarm}{HTML}{D62728}
\definecolor{wifigreen}{HTML}{00FF00}
\usepackage{ulem}
\usepackage{xfp}
\newcommand{\emphline}[1]{\uline{\textit{#1}}}
\def\onedot{.\xspace}
\def\eg{\emph{e.g}\onedot}

\definecolor{nvidiagreen}{RGB}{119,185,0}
\definecolor{pgonegreen}{RGB}{17,166,80}
\definecolor{stmmalarm}{HTML}{2CA02C}    
\definecolor{stmmdoorbell}{HTML}{D4AC0D} 
\definecolor{AudioMaxColor}{HTML}{4E81A9}
\definecolor{AudioMinColor}{HTML}{D3A53E}
\definecolor{TrainColor}{HTML}{4A90E2}
\definecolor{TestColor}{HTML}{7F8C8D}
\definecolor{room6}{HTML}{91A83B}
\definecolor{room7}{HTML}{625DA8}
\definecolor{room8}{HTML}{EF7565}
\definecolor{room16}{HTML}{984B56}
\definecolor{room17}{HTML}{E9A13B}
\newlength\savewidth\newcommand\shline{\noalign{\global\savewidth\arrayrulewidth
  \global\arrayrulewidth 1pt}\hline\noalign{\global\arrayrulewidth\savewidth}}
\definecolor{forestgreen}{RGB}{34, 139, 34}

\newcommand{\reffig}[1]{Figure~\ref{#1}}
\newcommand{\reftab}[1]{Table~\ref{#1}}

\newcommand{\Rnum}[1]{\uppercase\expandafter{\romannumeral #1\relax}}
\newcommand{\textitbf}[1]{\textbf{\textit{#1}}}
\newsavebox{\STMMMeasureBox}
\newsavebox{\STMMTableBox}

\newlength{\STMMPanelWidth}
\newlength{\STMMTopHeight}
\newlength{\STMMBottomHeight}
\newlength{\STMMTempHeight}
\newlength{\STMMTableHeightOne}
\newlength{\STMMTableHeightTwo}

\newcommand{\STMMMeasureHeight}[2]{%
  \sbox{\STMMMeasureBox}{#2}%
  \setlength{#1}{%
    \dimexpr\ht\STMMMeasureBox+\dp\STMMMeasureBox\relax
  }%
}
\newcommand{\STMMPlacePanel}[3]{%
  \makebox[\linewidth][c]{%
    \raisebox{#2}[#1][0pt]{%
      \raisebox{\depth}{#3}%
    }%
  }%
}

\title{MuSTMM: Learning Sound-Triggered Mobile Manipulation across Multiple Scenes}

\newlength{\AuthorCellWidth}

\newcommand{\AuthorCell}[2]{%
  \begin{minipage}[t]{\AuthorCellWidth}%
    \centering
    {\large #1\par}%
    \vspace{0.2em}%
    {\small #2\par}%
  \end{minipage}%
}

\newcommand{\AuthorRow}[2]{%
  \begingroup
  \setlength{\AuthorCellWidth}{\dimexpr .92\textwidth/#1\relax}%
  \parbox[c]{.92\textwidth}{%
    \noindent\makebox[.92\textwidth][c]{#2}%
  }%
  \endgroup
}
\author{\texorpdfstring{%
  \mdseries
  \begin{tabular}{@{}c@{}}
    \AuthorRow{4}{%
      \AuthorCell{Hao Ju\thanks{denotes equal contribution.}}{University of Macau}%
      \AuthorCell{Feng Yan$^*$}{Enacta-AI}%
      \AuthorCell{Zhihao Zhong}{Enacta-AI}%
      \AuthorCell{Heng Yang}{Enacta-AI}%
    }\\
    \noalign{\vskip 3ex}
    \AuthorRow{3}{%
      \AuthorCell{Shaofei Huang}{Hefei University of Technology}%
      \AuthorCell{Zhedong Zheng}{University of Macau}%
    }
  \end{tabular}%
}{Hao Ju, Feng Yan, Zhihao Zhong, Heng Yang, Shaofei Huang, Zhedong Zheng}}

\iclrfinalcopy
\begin{document}

\maketitle
\fancyhead{}
\fancyhead[L]{Under Review}
\begin{center}
  \vspace{-1em}
  \includegraphics[width=\textwidth]{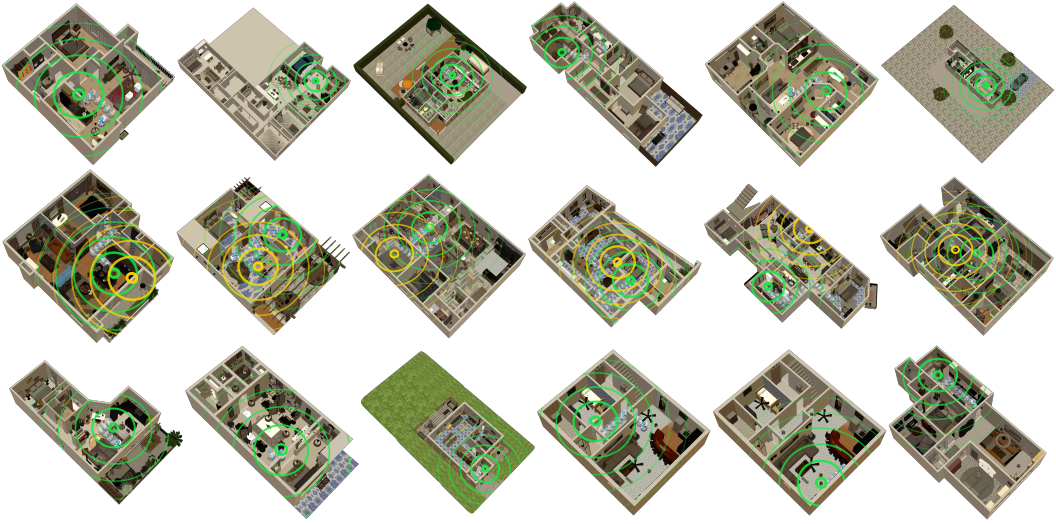}
  \vspace{-0.1in} 
  \captionof{figure}{
    {Example scenes in MuSTMM} (Zoom in for a better visualization).
    MuSTMM supports sound rendering (\textcolor{stmmalarm}{single} event and \textcolor{stmmalarm}{multi}\textcolor{stmmdoorbell}{ple} events) and physical interaction in 31 scenes, ranging from different architectural structures and household layouts.
    }
  \label{fig:motivation}
\end{center}
\vspace{0.1in} 

\begin{abstract}
Sound-Triggered Mobile Manipulation (STMM) requires robots to respond to sound events by navigating to the associated objects and performing appropriate physical interactions.
Existing STMM evaluations primarily vary object layouts within limited architectural structures, leaving challenges to unseen scenes insufficiently explored.
We introduce \textbf{Multi-scene Sound-Triggered Mobile Manipulation (MuSTMM)}, a dataset that combines spatial audio rendering and physical interaction across 31 scenes with diverse architectures and object layouts.
MuSTMM contains 14.36 million precomputed binaural Room Impulse Responses (RIRs) and a scene-disjoint split for evaluating single-event and multi-event settings in unseen scenes.
We further augment the Navigation Policy Model with explicit spatial memory that complements acoustic cues with observed geometry.
Specifically, we accumulate pose-aligned depth observations in a shared coordinate frame and project this memory into an egocentric Bird's-Eye View (BEV) occupancy map.
Fusing map features with sound context produces a goal descriptor that provides both target-location and scene-structure priors, guiding navigation to the target region required for subsequent physical interaction. This policy achieves a 3.81\% success rate boost over existing methods, while component ablations further support the effectiveness of spatial memory.
MuSTMM and the navigation policy provide a testbed for the new challenge of studying sound-triggered mobile manipulation in unseen scenes.
\textit{The code and dataset will be released upon acceptance.}
\end{abstract}

\section{Introduction}
\label{sec:introduction}

Sound events provide a natural trigger for robots to perceive their surroundings and interact with physical objects.
Sound-Triggered Mobile Manipulation (STMM)~\citep{ju_stmm_arxiv26} studies this capability by requiring a robot to respond to sound events, navigate to the associated objects, and perform appropriate physical interactions.
These interactions include relocating sounding rigid objects and changing the states of articulated objects, either for a single event or sequentially for multiple events.
Existing benchmarks provide only partial challenges of different scene structures and object layouts.
Audio-visual navigation benchmarks, such as SoundSpaces~\citep{chen_soundspaces_eccv20} and SAVi~\citep{chen_savi_cvpr21}, support sound rendering across multiple scenes but do not include the physical interactions required by STMM.
Conversely, mobile manipulation benchmarks, such as Habitat-M3~\citep{gu_habmm_iclr23} and ManiSkill-Hab~\citep{shukla_maniskill_iclr25}, support physical interaction in a limited structure variation without sound rendering.
STMM~\citep{ju_stmm_arxiv26} brings these capabilities together, but evaluates object-layout variation within limited scene structures.
Although varying object placements brings partial layout variation, it does not evaluate whether agents can perceive and respond to sound events in unseen structures.
This motivates a benchmark that combines structural diversity with both sound rendering and physical interaction.

To address this gap, we introduce \textbf{MuSTMM}, a multi-scene benchmark that supports sound rendering and physical interaction across 31 scenes from HSSD~\citep{khanna2024hssd} (see~\reffig{fig:motivation}).
MuSTMM expands variation at two complementary levels.
\textit{Structure variation} covers differences in scene architecture, including room geometry and spatial connectivity.
\textit{Layout variation} combines episode-specific placements of sounding targets and silent source-capable distractors with scene-specific arrangements of non-source background furniture.
To support sound rendering across these scenes, we leverage SoundSpaces~\citep{chen2022soundspaces2} to densely precompute 14.36 million binaural Room Impulse Responses (RIRs), providing scene-specific sound rendering.
A scene-disjoint split, with 23 training scenes and 8 held-out test scenes, enables evaluation of both single-event and multi-event tasks in unseen scenes.
The accompanying configurable simulator and data-generation pipeline also supports extensions with additional scenes, assets, and task configurations.

We study MuSTMM with a hierarchical framework in which a Sound-triggered Task Planner uses the initial RGB and audio observations to predict an ordered skill chain, which is executed by the corresponding skill-specific policy models.
Among the low-level policy models, sound navigation policy brings the robot to the target region required for subsequent interaction, providing the spatial precondition for executing the next skill.
This cross-scene setting motivates a closer examination of how navigation policies estimate and represent sound-source goals.
Binaural sound supplies directional cues about the source but contains limited geometry information of the observed scene.
Instead, vision observations contain rich geometry priors.
However, this geometric context is distributed across observations acquired at different viewpoints and time steps.
Previous methods encode vision and sound observations into temporal memories~\citep{zeng2026savnce,chen_savi_cvpr21}. But the geometry information of the scene may misalign across different time steps in this way.
To this end, we augment memory-based goal estimation with explicit spatial memory.
Rather than relying on compressed observation history alone to represent observed geometry, we accumulate depth observations in a shared coordinate frame, and project them into an egocentric BEV occupancy map.
The map records occupancy and exploration status, while its features are fused with acoustic and temporal context to form a goal descriptor.
In this way, the explicit spatial memory contains both goal-related cues and observed scene structure, providing priors for the navigation policy.

We evaluate MuSTMM under a scene-disjoint protocol in single-event and multi-event settings.
Results identify navigation as a major execution bottleneck in held-out scenes, even when ground-truth skill chains are provided.
Navigation Policy Model with explicit spatial memory achieves a navigation success rate of 75.28\%, outperforming existing methods.
Memory ablations further show that combining explicit spatial memory with compressed spatio-temporal memory outperforms either component alone, highlighting their complementary roles in across-scene sound navigation.

Our contributions are threefold:
\begin{itemize}[leftmargin=0pt]
    \item We introduce \textbf{MuSTMM}, a multi-scene benchmark that combines spatial audio rendering and physical interaction with architectural structure variation and object layout variation. MuSTMM consists of 30.11k episodes. The scene-disjoint evaluation protocol and configurable data-generation pipeline support the study of sound-triggered mobile manipulation in unseen scenes.
    \item We augment the Goal Descriptor Branch of the {Navigation Policy Model} with \textbf{explicit spatial memory} accumulated from depth observations. By integrating observed geometry with sound context, the resulting goal descriptor provides target-related and structural priors for the navigation policy model and ensures subsequent interaction. 
    \item We systematically evaluate baselines in held-out scenes under both single-event and multi-event settings. Navigation comparisons and memory ablations show that explicit spatial memory complements compressed observation history to improve sound-source navigation. Despite these gains, results of the Oracle baseline identify navigation as a major bottleneck in unseen scenes, highlighting a key direction for future research.
\end{itemize}

\section{Related Work}
\label{sec:related_work}

\noindent\textbf{Embodied Navigation and Manipulation Benchmark.}
Audio-visual navigation and mobile manipulation benchmarks provide complementary capabilities for embodied agents.
SoundSpaces~\citep{chen_soundspaces_eccv20,chen2022soundspaces2} support spatial audio rendering across diverse indoor scenes, while SAVi~\citep{chen_savi_cvpr21} introduces semantic sound events for target navigation.
These benchmarks require agents to reach sound sources but do not include the subsequent physical interactions.
In parallel, Habitat-M3~\citep{gu_habmm_iclr23} studies mobile manipulation for household rearrangement, while ManiSkill-HAB~\citep{shukla_maniskill_iclr25} emphasizes GPU parallelism in home rearrangement tasks.
Their ReplicaCAD-based evaluations vary furniture layouts and object placements within a shared architectural structure, providing {layout variation} but limited {structure variation}.
HomeRobot~\citep{yenamandra2023homerobot} evaluates open-vocabulary mobile manipulation across diverse indoor scenes, with scene-disjoint splits to assess generalization to unseen scenes.
However, these mobile manipulation benchmarks do not incorporate spatial sound rendering.
STMM~\citep{ju_stmm_arxiv26} combines sound rendering and physical interaction through Habitat-Echo, requiring robots to respond to sound events by navigating to and manipulating the associated objects.
However, the evaluation primarily varies object configurations within limited architectural structures, leaving performance in unseen scenes under-explored.
The proposed MuSTMM introduces architecturally distinct scenes together with object-layout variation, preserving both sound rendering and physical interaction at the same time.
The scene-disjoint evaluation protocol supports the study of sound-triggered mobile manipulation in unseen scenes.

\noindent\textbf{Embodied Navigation.}
Explicit spatial representations have long been used to accumulate observations and support planning in visual navigation~\citep{chaplot2020learning,gupta2017cognitive,ramakrishnan2020occupancy}.
Trans4Map~\citep{chen2023trans4map} aggregates egocentric visual features into allocentric memory for Bird's-Eye-View (BEV) semantic mapping.
GridMM~\citep{wang2023gridmm} and BSG~\citep{liu2023bird} organize historical observations into grid-based memory and BEV scene graphs, respectively, supporting spatial reasoning and navigation decisions.
VLN-Game~\citep{yu2026vlngame} uses object-centric spatial maps for language-guided target search, while NavA$^3$~\citep{zhang2026nava3} exploits top-down scene representations for target-region reasoning and navigation.
Audio-visual navigation similarly benefits from structured spatial information, but requires agents to infer goals from acoustic observations~\citep{gan2020look}.
AV-WaN~\citep{chen2020learning} incorporates spatially grounded acoustic memory into waypoint-based navigation.
SAVi~\citep{chen_savi_cvpr21} introduces a goal descriptor capturing the target's spatial and semantic properties, together with persistent multimodal memory.
MAGNet~\citep{zeng2026savnce} further combines auditory cues, self-motion information, and episodic memory to maintain goal estimates over time.
Our method complements memory-based goal estimation with explicit geometry accumulated from pose-aligned depth observations.
The goal descriptor provides both target-related and structural priors for the policy branch, supporting navigation to the target region required for subsequent interaction.

\begin{figure}[t]
    \vspace{-2em}
    \centering

    \makebox[\linewidth][c]{%
    \begin{subfigure}[c]{0.6\linewidth}
        \raggedleft
        \small
        \setlength{\tabcolsep}{3pt}
        \renewcommand{\arraystretch}{0.95}

        \resizebox{\linewidth}{!}{%
        \begin{tabular}{ccccc}
            \shline
            \multirow{2}{*}{Dataset}
            & Physical
            & Sound
            & \multicolumn{2}{c}{Variation} \\
            \cmidrule(lr){4-5}
            & Interaction
            & Rendering
            & Structure
            & Layout \\
            \hline

            SoundSpaces~\citep{chen_soundspaces_eccv20}
            & \textcolor{red}{\ding{55}}
            & \textcolor{nvidiagreen}{\ding{51}}
            & \textcolor{nvidiagreen}{\ding{51}}
            & \textcolor{red}{\ding{55}} \\

            SAVi~\citep{chen_savi_cvpr21}
            & \textcolor{red}{\ding{55}}
            & \textcolor{nvidiagreen}{\ding{51}}
            & \textcolor{nvidiagreen}{\ding{51}}
            & \textcolor{red}{\ding{55}} \\

            Habitat-M3~\citep{gu_habmm_iclr23}
            & \textcolor{nvidiagreen}{\ding{51}}
            & \textcolor{red}{\ding{55}}
            & \textcolor{red}{\ding{55}}
            & \textcolor{nvidiagreen}{\ding{51}} \\

            ManiSkill-Hab~\citep{shukla_maniskill_iclr25}
            & \textcolor{nvidiagreen}{\ding{51}}
            & \textcolor{red}{\ding{55}}
            & \textcolor{red}{\ding{55}}
            & \textcolor{nvidiagreen}{\ding{51}} \\

            STMM~\citep{ju_stmm_arxiv26}
            & \textcolor{nvidiagreen}{\ding{51}}
            & \textcolor{nvidiagreen}{\ding{51}}
            & \textcolor{red}{\ding{55}}
            & \textcolor{nvidiagreen}{\ding{51}} \\

            \hline
            MuSTMM (\textbf{Ours})
            & \textcolor{nvidiagreen}{\ding{51}}
            & \textcolor{nvidiagreen}{\ding{51}}
            & \textcolor{nvidiagreen}{\ding{51}}
            & \textcolor{nvidiagreen}{\ding{51}} \\

            \shline
        \end{tabular}
        }

        \caption{}
        \label{fig:dataset_comparison_table}
    \end{subfigure}
    \hspace{0.015\linewidth}
    %
    \begin{subfigure}[c]{0.2\linewidth}
        \raggedright
        \includegraphics[width=\linewidth]{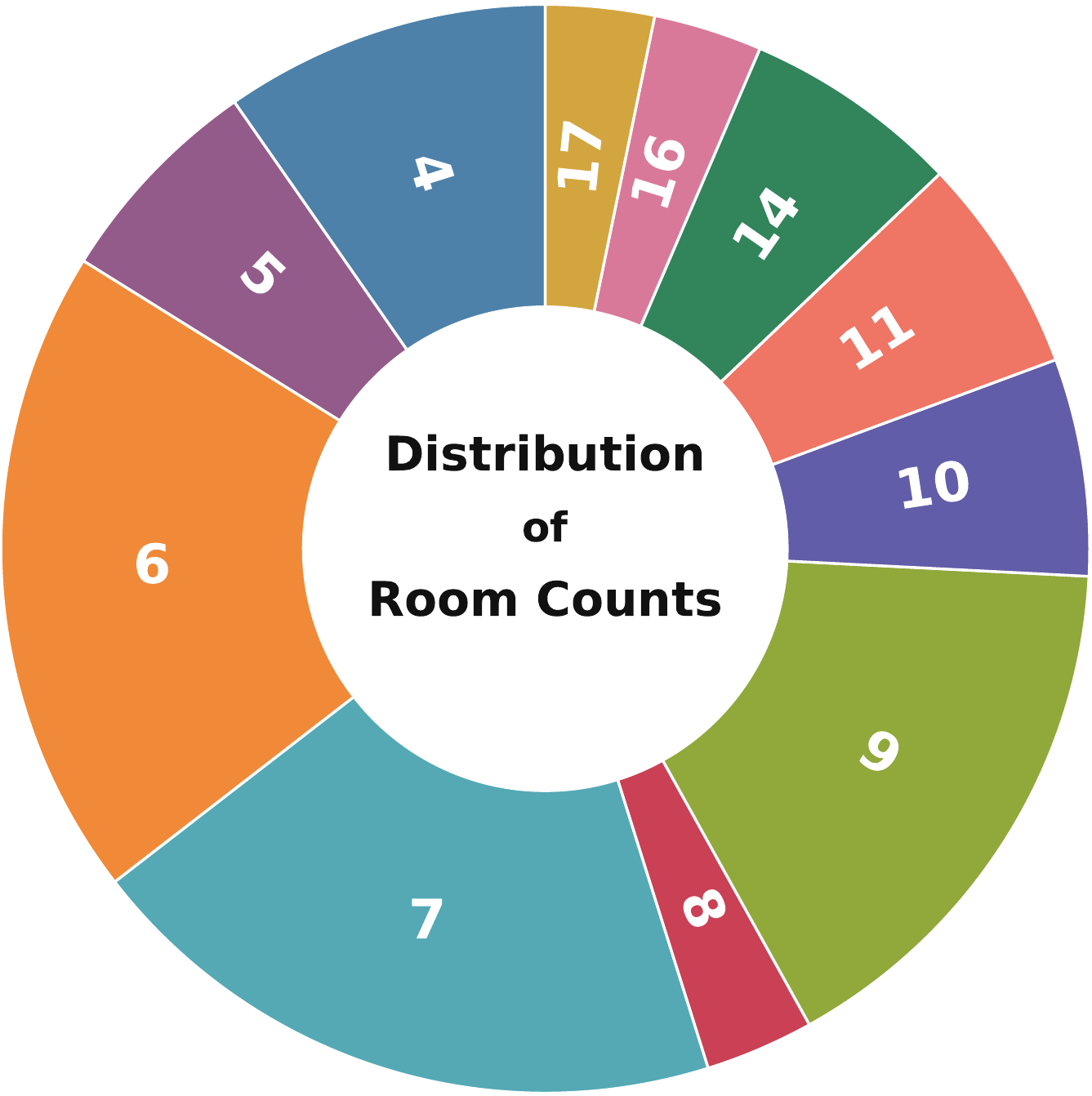}

        \caption{}
        \label{fig:dataset_visualization}
    \end{subfigure}
    }

    \par\vspace{2pt}
    \makebox[\linewidth][c]{%
        \begin{subfigure}[c]{0.855\linewidth}
            \centering
            \includegraphics[width=\linewidth]{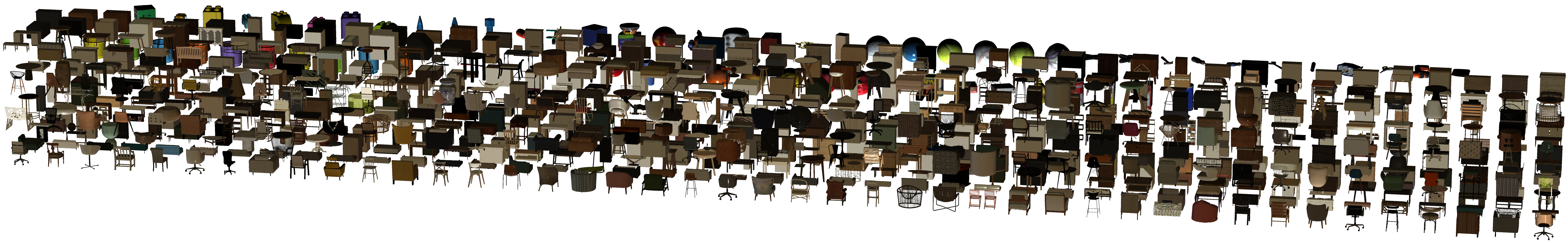}
            \vspace{-2em}
            \caption{}
            \label{fig:furniture}
        \end{subfigure}
    }
    \vspace{-1em}
    \caption{
    {Overview of MuSTMM.}
    {(a)~Comparison with existing datasets}. In addition to physical interaction and sound rendering, MuSTMM brings structure and layout variation.
    {(b)~Structure variation}: Distribution of room counts across the different scenes in MuSTMM, up to \textcolor{room17}{17}.
    {(c)~Layout variation}: MuSTMM contains 855 furniture assets, providing varied visual backgrounds.
    }
    \label{fig:dataset}
    \vspace{-1.5em}
\end{figure}

\section{The MuSTMM dataset}
Sound-Triggered Mobile Manipulation (STMM)~\citep{ju_stmm_arxiv26} requires a robot to respond to sound events by navigating to the sound source objects and performing the appropriate physical interactions.
In the single-event setting, the robot relocates sounding rigid objects or transitions the state of articulated objects. In the multi-event setting, the robot interacts with two sounding objects sequentially according to a given priority.
Supporting these tasks requires both spatial audio rendering and physical interaction. MuSTMM provides these capabilities across 31 scenes (see~\reffig{fig:dataset_comparison_table}), with additional variation at two complementary levels: architectural structure across scenes (structure variation) and object layout within and across scenes (layout variation).

\noindent\textbf{Structure Variation.}
Structure variation refers to differences in scene architecture, including the number, shape, and size of rooms, spatial connectivity, geometry of architectural elements, and so on.
We introduce 31 scenes from HSSD~\citep{khanna2024hssd} for STMM, configuring each physical interactive scene to support acoustic rendering.
As one indicator of structural diversity, the introduced scenes cover 11 distinct room counts, ranging from four-room scenes to scenes with 17 rooms
(see~\reffig{fig:dataset_visualization}).
For sound rendering across these scenes, we use SoundSpaces~\citep{chen2022soundspaces2} to densely precompute binaural Room Impulse Responses (RIRs) through bidirectional acoustic path tracing.
We sample source-receiver position pairs and consider four receiver orientations, obtaining 14.36 million RIRs across the 31 scenes.
The resulting collection provides scene-specific acoustic responses, supporting sound rendering and physical interaction at the same time.

\noindent\textbf{Layout Variation.}
Layout variation concerns the spatial configuration of objects relative to the architectural structure at the scene level.
During episode generation, we insert sounding targets and silent source-capable objects at randomly sampled locations. Source-capable objects can serve as sound sources but remain silent in the current episode, acting as instance-level visual distractors.
This episode-wise placement allows each scene to support multiple configurations of sounding targets and silent distractors.
Complementing these task-related objects, MuSTMM includes 855 non-sound-source household furniture assets, spanning categories such as chairs and cabinets (see~\reffig{fig:furniture}). These background furnishings are distinct from both the sounding targets and the silent source-capable distractors. Their scene-specific arrangements provide diverse visual backgrounds and spatial contexts for task execution.
MuSTMM thus combines episode-level source-object placement variation with diverse non-source furnishings across scenes.

\noindent\textbf{Episode Generation and Dataset Splits.}
By episode-wise object placements, we construct 24.91k training episodes across 23 scenes and 5.20k testing episodes across 8 held-out scenes, covering both single-event and multi-event settings.
The training and testing scene sets are disjoint.

\noindent\textbf{Discussion.~The contribution to the community.}
\emphline{(1)~Unseen scene-level challenge}.
The scene-disjoint split provides a testbed for assessing whether agents can ground sound events and perform the required physical interactions in unseen scenes.
\emphline{(2)~Open-source data generator}.
Beyond the cross-scene challenge, the configurable simulator and data-generation pipeline allows researchers to extend the dataset with additional scenes, assets, and task configurations as needed.

\begin{figure*}[!t]
    \centering
    \includegraphics[width=0.98\linewidth]{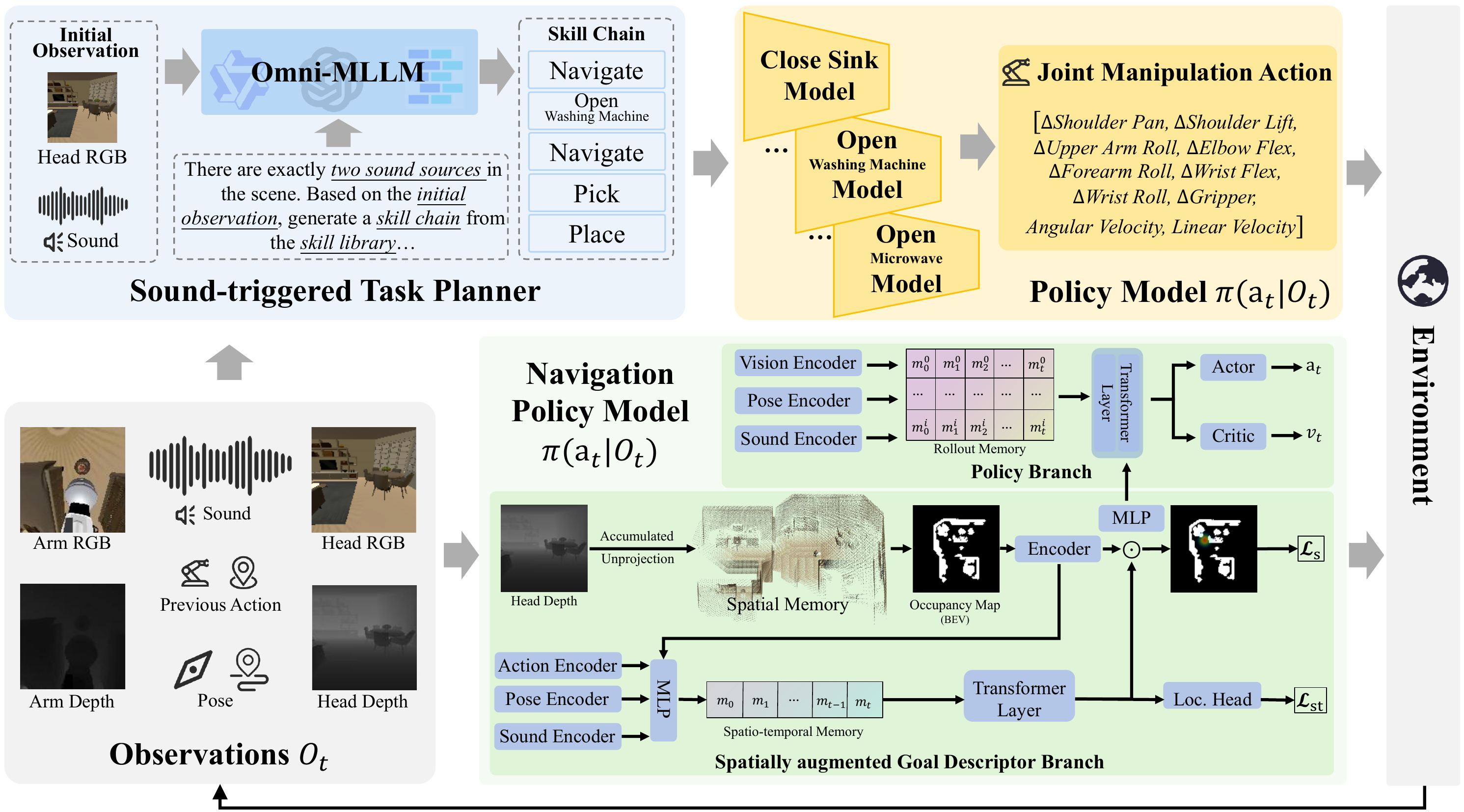}
    \vspace{-.5em}
    \caption{
    {Overview of the proposed baseline.}
    The baseline consists of a high-level task planner and low-level policy models.
    {Sound-triggered Task Planner} receives initial observations and predicts the skill chain.
    Following the skill chain, {policy models} are sequentially activated and interact with the environment.
    For the {navigation policy model}, we propose spatially augmented goal descriptor branch that provides the policy branch with spatio-temporal context of the goal.
    (Color information is projected into the spatial memory for visualization only.)
    }
    \vspace{-1em}
    \label{fig:method}
\end{figure*}

\section{Method}
\label{sec:method}
As illustrated in~\reffig{fig:method}, our framework consists of a Sound-triggered Task Planner and a set of learned, skill-specific policy models. The high-level planner predicts an ordered skill chain, and the corresponding low-level policies execute sequentially.
Within the low-level skill chain, the navigation policy brings the robot to the target region required for subsequent interaction, enabling progression through successive skills. We incorporate spatial memory alongside sound context to augment the goal description in the navigation policy model.

\subsection{Task Planning and Policy Execution}
\label{sec:task_planning}
\noindent\textbf{Sound-triggered Task Planner.}
Given a predefined skill library, the planner selects and orders the skills required to address the observed sound events from sound and vision observations. The planner is invoked only once at the beginning of each episode without additional replanning. In the multi-event setting, the planner receives an additional priority of events.

\noindent\textbf{Policy Model.}
Corresponding policy models are activated sequentially based on the skill chain.
Each policy receives observations to predict control actions. 
For interaction skills involving articulated objects, the policy jointly outputs the base linear and angular velocities, seven arm-joint actions, and one gripper action, coordinating base and arm motion at each control step.

\subsection{Navigation Policy Model}
\label{sec:navigate_policy}
The navigation policy model connects task planner and different policies by bringing the robot to the target region required for subsequent execution, which makes navigation essential among policy models.
%
Navigation Policy Model comprises a Spatially augmented Goal Descriptor Branch that predicts the location of the sound source and a Policy Branch that outputs navigation actions from the predicted goal descriptor and observation history. We propose to accumulate visual geometry that serves as spatial memory in the Goal Descriptor Branch (see~\reffig{fig:method}).

\noindent\textbf{Spatially augmented Goal Descriptor Branch.}
At each step, we unproject the depth observation using camera intrinsics and transform the resulting points into world coordinates using camera extrinsics. These points incrementally update a persistent point cloud, forming a spatial memory of the past and current observations. Using the current camera pose, we project this memory into an egocentric Bird's-Eye View (BEV) and construct a two-channel occupancy map. The first channel records occupancy, while the second records exploration status, distinguishing unobserved regions from observed free space.
The occupancy map is processed by a CoordConv-based encoder~\citep{liu2018intriguing} and fused with encoded audio, pose, and previous-action features. These representations compose the compressed spatio-temporal memory~\citep{fang2019scene} from previous and current steps. Whereas the spatial memory stores accumulated geometry, the spatio-temporal memory retains the history of multimodal features used for goal estimation. Transformer layers process the spatio-temporal memory to produce a spatio-temporal embedding, which is passed to a location head to predict the sound-source position.
Alongside the location head, we fuse the spatio-temporal embedding with the occupancy map feature to predict the sound-source location. The fused map features provide the goal and overall geometry information for the Policy Branch.

\noindent\textbf{Policy Branch.}
We adopt a transformer encoder-decoder structure following prior navigation policies~\citep{chen_savi_cvpr21,zeng2026savnce}. Each observation is processed by the corresponding encoder, and the fused observation features are appended to the rollout memory $m_t^i$ across different environments. A transformer encoder processes this history, providing keys and values for the decoder. Meanwhile, the fused map features from the Goal Descriptor Branch are projected through an MLP to form the decoder query. The decoder output is passed to actor and critic heads for action prediction and value estimation, respectively.

\noindent\textbf{Discussion. What are the advantages of spatially augmented goal descriptor branch?}
\emphline{(1)~Vision observation is complementary to sound for navigation policy.}
Binaural sound contains directional cues about the sound source but limited geometric clues, while accumulated depth observations provide geometric information of the observed scene.
The fusion of vision and sound observations supplies geometric context for interpreting directional cues and estimating the sound-source location.
\emphline{(2)~Vision observations are explicit spatial memory for navigation policy.}
Given observations at different timesteps, previous method~\citep{zeng2026savnce} encodes vision observations into spatio-temporal memory as other observations, highlighting the temporal consistency within vision information.
The proposed spatially augmented goal descriptor explicitly preserves spatial information and accumulates it across different timesteps, forming the spatial memory over previous observations. 
The occupancy map further expresses this geometry relative to the robot's current pose.
The policy model does not need to establish correspondences between complicated spatio-temporal memories and current observations.
\emphline{(3)~Occupancy map feature is a prior of both sound source and scene structure.}
Spatial memory and occupancy map provide an overall structure prior for the scene, which has been proven effective in VLN work~\citep{chen2023trans4map,yu2026vlngame,zhang2026nava3}.
Occupancy map feature in the proposed spatially augmented goal descriptor contains priors of goal location and scene-level structure priors at the same time.
The goal prior is from the fusion of occupancy map feature and spatio-temporal embedding (especially binaural sound).
Occupancy map contains passable paths and obstacles, providing the structure prior.
With the help of the goal descriptor branch, the policy branch can be aware of where (goal prior) and how (structure prior) to navigate toward the sound source.

\subsection{Learning objective}
We train policy models using PPO~\citep{schulman_ppo_arxiv17}. Below we take the navigation policy model as an example. Others are similar.

\noindent\textbf{Reward for Navigation Policy Model.}
We train the navigation policy model using a dense progress reward, with additional penalties for time and collisions. At step $t$, the reward is defined as
\begin{equation}
\begin{aligned}
r_t ={}& -0.001
    + \left(d_{t-1}-d_t\right)
    -0.01\,\mathbb{I}_{\mathrm{invalid},t}
    + 2.5\mathbb{I}_{\mathrm{succ},t},
  \label{eq:nav_reward}
  \end{aligned}
  \end{equation}
where $-0.001$ is a slack penalty applied at each step. 
$d_t$ denotes the shortest collision-safe geodesic distance from the current position to the nearest valid navigation goal at timestep $t$, such that progress toward the goal yields a positive reward and movement away from it incurs a penalty. 
$\mathbb{I}_{\mathrm{invalid},t}$ denotes unsafe actions that increase robot collision force, with the corresponding penalty capped at $0.01$ per step.
$\mathbb{I}_{\mathrm{succ},t}$ is the success indicator, which is activated only when the policy stops within the goal region.

\noindent\textbf{PPO Loss for the Policy Branch.}
Given the dense reward $r_t$, we compute the return target $\hat{R}_t$ using Generalized Advantage Estimation (GAE)~\citep{schulman2015high} and define the advantage as
\begin{equation}
\hat{A}_t = \hat{R}_t - V(x_t),
\end{equation}
where $x_t$ is the policy input and $V(x_t)$ is the critic prediction.

The PPO loss is defined as
\begin{equation}
\begin{aligned}
\mathcal{L}_{\mathrm{PPO}}
={}&
-\mathbb{E}_{t}\left[
\min\left(
\rho_t\hat{A}_t,\,
\operatorname{clip}(\rho_t,1-\epsilon,1+\epsilon)\hat{A}_t
\right)
\right] \\
&+ c_v\,\mathbb{E}_{t}\left[
\left(V_{\theta}(x_t)-\hat{R}_t\right)^2
\right]
-\beta\,\mathbb{E}_{t}\left[
\mathcal{H}\bigl(\pi_{\theta}(\cdot\mid x_t)\bigr)
\right],
\label{eq:ppo_loss}
\end{aligned}
\end{equation}
where
\begin{equation}
\rho_t =
\frac{\pi_{\theta}(a_t\mid x_t)}
{\pi_{\theta_{\mathrm{old}}}(a_t\mid x_t)}.
\end{equation}

Here, \(\rho_t\) denotes the probability ratio between the current and behavior policies, and \(\hat{A}_t\) is the normalized advantage estimated using GAE.
The first term improves the policy by increasing the probability of actions with positive advantages while limiting overly large policy updates.
The second term trains the critic to predict the return target \(\hat{R}_t\). The entropy term \(\mathcal{H}\) encourages exploration by preventing the policy from becoming prematurely deterministic.

\noindent\textbf{Loss for the Spatially augmented Goal Descriptor Branch.}
This branch estimates the target location from binaural audio, occupancy map, and other observations.
For each valid rollout step \(t\), the ground-truth sound-source location \(\mathbf{y}_t\) is represented in the robot base frame. We train the goal predictor using the following loss:
\begin{equation}
\begin{aligned}
\mathcal{L}_{\mathrm{loc}}
= {}&\operatorname{\mathcal{L}_{st}}\! + \operatorname{\mathcal{L}_s}\!
={}&
\operatorname{\mathcal{L}_1}\!\left(
\hat{\mathbf{u}}_t,\mathbf{u}_t
\right)
+
\operatorname{\mathcal{L}_1}\!\left(
\hat{s}_t,s_t
\right)
+
\operatorname{\mathcal{L}_{ce}}\!\left(
P_t,G_t
\right)
+
\operatorname{\mathcal{L}_{bce}}\!\left(
Z_t,G_t
\right)
+
\operatorname{\mathcal{L}_1}\!\left(
\hat{\mathbf{y}}_t,\mathbf{y}_t
\right),
\label{eq:location_prediction_loss}
\end{aligned}
\end{equation}
where \(\mathcal{L}_{1}\), \(\mathcal{L}_{\mathrm{CE}}\), and
\(\mathcal{L}_{\mathrm{BCE}}\) denote the Smooth \(L_1\), cross-entropy, and
binary cross-entropy losses, respectively.
The first two terms form \(\mathcal{L}_{st}\) which supervise the predicted target direction
\(\hat{\mathbf{u}}_t\) and normalized distance \(\hat{s}_t\) based on the spatio-temporal memory.
The remaining three terms are \(\mathcal{L}_{s}\) on the spatial memory.
\(Z_t\) denotes the predicted location logits on the occupancy map, \(P_t\) is the corresponding spatial softmax distribution, and \(G_t\) is a Gaussian target heatmap centered
at the ground-truth location.
\(\mathcal{L}_{\mathrm{ce}}\) and \(\mathcal{L}_{\mathrm{bce}}\) aligns the predicted spatial distribution over the occupancy map.
The final Smooth \(L_1\) term directly aligns the soft-argmax location \(\hat{\mathbf{y}}_t\) with the ground-truth in Cartesian coordinates.
The location predictor is optimized separately from the PPO policy.

\section{Experiment}
\noindent\textbf{Implementation Detail.} 
Policies are trained via PPO~\citep{schulman_ppo_arxiv17} on an NVIDIA A100 GPU for $5 \times 10^6$ environment steps. Optimization relies on Adam~\citep{kingma_adam_arxiv14} with an initial learning rate of $1 \times 10^{-4}$ (linear decay), a weight decay of $1 \times 10^{-5}$.

\noindent\textbf{Baseline.} 
We combine individually trained low-level policies with three types of task planners, composing three baselines.
(1) \textbf{Oracle}: Uses ground-truth skill chains, serving as an upper bound of task planning.
(2) \textbf{Zero-shot Omni-MLLMs}: Evaluates training-free planners including Qwen2.5-Omni (3B, 7B)~\citep{xu_qwen2_arxiv25}, MiniCPM-o-4.5~\citep{cui_minicpm_arxiv26}, and Qwen3-Omni-30B-Instruct~\citep{yang_qwen3_arxiv25}.
(3) \textbf{Supervised Setting-Specific Planner (SSP)}: A lightweight network is trained to map initial observations to skill chains via sound category prediction. It processes log-Mel sound and head RGB for the single-event setting, and priority embeddings for the multi-event setting.

\noindent\textbf{Evaluation Protocol.}
Following previous work~\citep{ju_stmm_arxiv26}, we evaluate baselines across various scenes in two settings. (1) \textit{Single-event}, which involves relocating rigid sound sources (object relocation) or changing the states of articulated ones (state transition). (2) \textit{Multi-event}, which requires the robot to sequentially interact with two sound sources. 
Performance is measured using two metrics. \textit{Planning Success Rate (PSR)} evaluates whether the task planner correctly predicts the skill chain. \textit{Completion Success Rate (CSR)} is a stricter metric, counting an episode as successful only if the plan is correct and all constituent skills are successfully executed in order.

\subsection{Result}

\begin{figure*}[!t]
\vspace{-1em}
  \centering

  \captionsetup[subfigure]{skip=3pt}

  \setlength{\STMMPanelWidth}{0.49\textwidth}
  \setlength{\tabcolsep}{3pt}

  \newcommand{\STMMResultsTable}{%
    \begin{tabular}{ccccccc}
    \shline
    \multirow{3}{*}[-1ex]{Method} & \multicolumn{4}{c}{Single-event setting} & \multicolumn{2}{c}{\multirow{2}{*}[-.5ex]{Multi-event setting}} \\
    \cmidrule(lr){2-5}
    & \multicolumn{2}{c}{Object Relocation} & \multicolumn{2}{c}{State Transition} & \multicolumn{2}{c}{} \\
    \cmidrule(lr){2-3} \cmidrule(lr){4-5} \cmidrule(lr){6-7}
    & Planning ($\uparrow$) & Completion ($\uparrow$) & Planning ($\uparrow$) & Completion ($\uparrow$) & Planning ($\uparrow$) & Completion ($\uparrow$) \\
    \hline
    Oracle & 100.00 & 66.67 & 100.00 & 63.12 & 100.00 & 39.50 \\
    \hline
    \hline
    Qwen2.5-Omni-3B & 0.36 & 0.30 & 18.44 & 12.56 & 65.90 & 28.30 \\
    Qwen2.5-Omni-7B & 5.11 & 2.61 & 32.31 & 22.34 & \textbf{76.10} & \textbf{31.60} \\
    MiniCPM-o-4.5 & 11.76 & 7.49 & 30.33 & 20.27 & 33.80 & 14.70 \\
    Qwen3-Omni-30B & \textitbf{63.76} & \textitbf{41.83} & \textitbf{38.35} & \textitbf{26.43} & 67.20 & 26.60 \\
    SSP & \textbf{94.00} & \textbf{62.51} & \textbf{61.49} & \textbf{39.51} & \textitbf{75.40} & \textitbf{30.60} \\
    \shline
    \end{tabular}
  }
  %
  %

  \def\STMMAHeightScale{0.80}
  \def\STMMAShift{10pt}

  \def\STMMBHeightScale{1.00}
  \def\STMMBShift{2pt}

  \def\STMMCHeightScale{0.95}
  \def\STMMCShift{4pt}

  \def\STMMDHeightScale{1.00}
  \def\STMMDShift{0pt}


  \STMMMeasureHeight{\STMMTopHeight}{%
    \includegraphics[
      width=\STMMPanelWidth
    ]{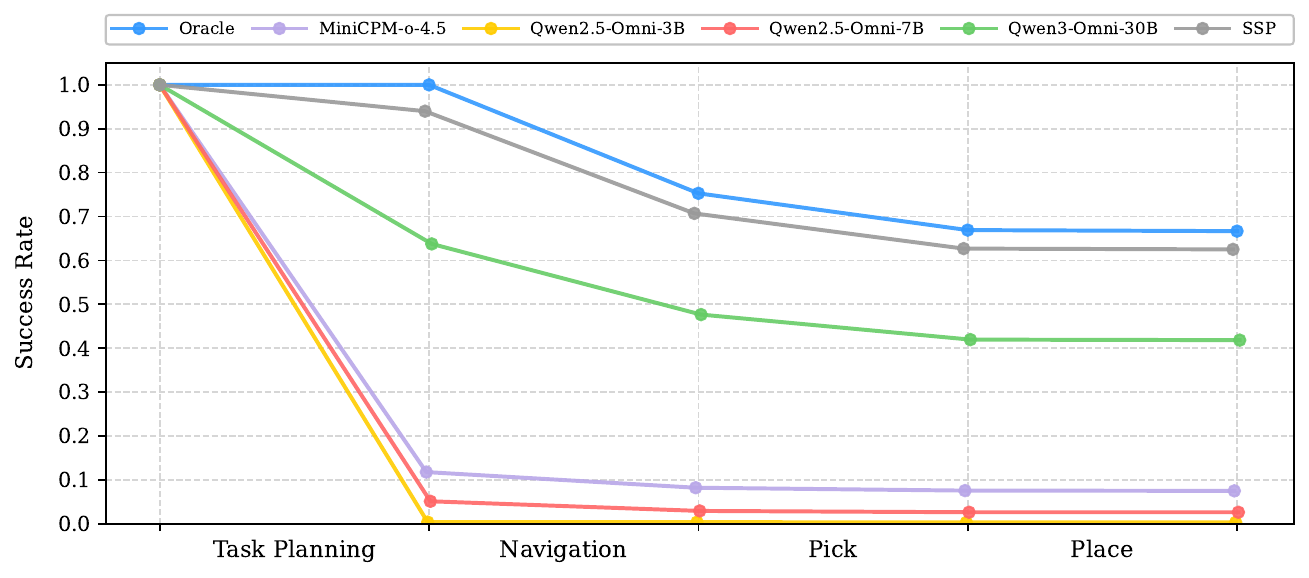}%
  }

  \STMMMeasureHeight{\STMMBottomHeight}{%
    \includegraphics[
      width=\STMMPanelWidth
    ]{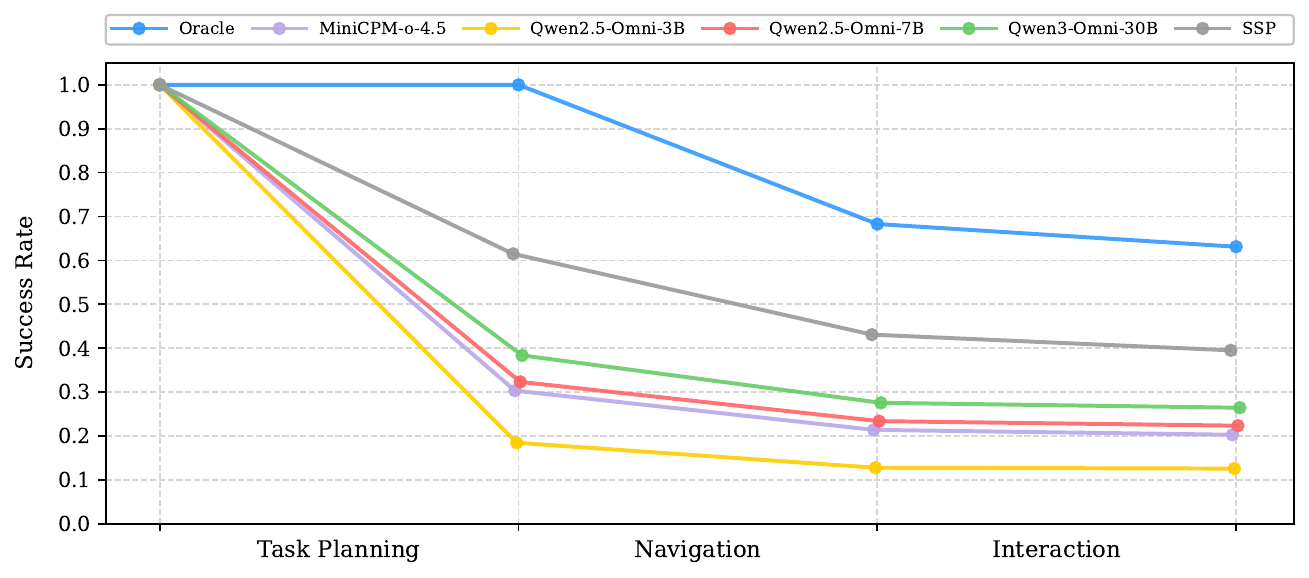}%
  }

  \STMMMeasureHeight{\STMMTempHeight}{%
    \includegraphics[
      width=\STMMPanelWidth
    ]{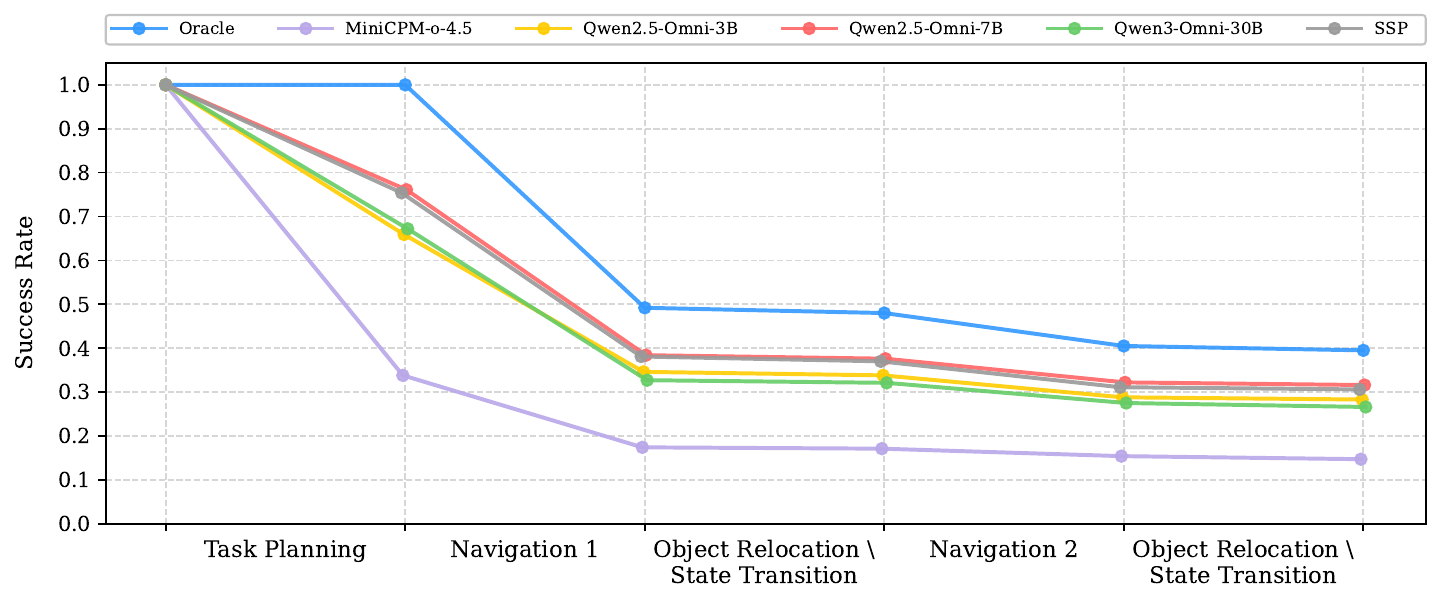}%
  }

  \ifdim\STMMTempHeight<\STMMBottomHeight
    \setlength{\STMMBottomHeight}{\STMMTempHeight}
  \fi


  \renewcommand{\arraystretch}{1}
  \STMMMeasureHeight{\STMMTableHeightOne}{%
    \resizebox{\STMMPanelWidth}{!}{\STMMResultsTable}%
  }

  \renewcommand{\arraystretch}{2}
  \STMMMeasureHeight{\STMMTableHeightTwo}{%
    \resizebox{\STMMPanelWidth}{!}{\STMMResultsTable}%
  }

  \edef\STMMTableStretch{\fpeval{%
    max(
      0.6,
      1 + (
        \STMMAHeightScale * \STMMTopHeight
        - \STMMTableHeightOne
      ) / (
        \STMMTableHeightTwo - \STMMTableHeightOne
      )
    )%
  }}

  \renewcommand{\arraystretch}{\STMMTableStretch}

  \sbox{\STMMTableBox}{%
    \resizebox{\STMMPanelWidth}{!}{\STMMResultsTable}%
  }


  \begin{subfigure}[b]{\STMMPanelWidth}
    \centering
    \STMMPlacePanel{\STMMTopHeight}{\STMMAShift}{%
      \usebox{\STMMTableBox}%
    }%
    \subcaption{}
    \label{fig:sota_overall}
  \end{subfigure}\hfill
  \begin{subfigure}[b]{\STMMPanelWidth}
    \centering
    \STMMPlacePanel{\STMMTopHeight}{\STMMBShift}{%
      \includegraphics[
        height=\STMMBHeightScale\STMMTopHeight
      ]{fig/success_rate_curve_relocation.pdf}%
    }%
    \subcaption{}
    \label{fig:pick_place}
  \end{subfigure}

  \par\vspace{1ex}


  \begin{subfigure}[b]{\STMMPanelWidth}
    \centering
    \STMMPlacePanel{\STMMBottomHeight}{\STMMCShift}{%
      \includegraphics[
        height=\STMMCHeightScale\STMMBottomHeight
      ]{fig/success_rate_curve_state_transition.pdf}%
    }%
    \subcaption{}
    \label{fig:state_transition}
  \end{subfigure}\hfill
  \begin{subfigure}[b]{\STMMPanelWidth}
    \centering
    \STMMPlacePanel{\STMMBottomHeight}{\STMMDShift}{%
      \includegraphics[
        height=\STMMDHeightScale\STMMBottomHeight
      ]{fig/success_rate_curve_multi_event.pdf}%
    }%
    \subcaption{}
    \label{fig:2sound}
  \end{subfigure}
  \vspace{-.8em}
  \caption{
  Results on MuSTMM. 
  (a) Success rates of task planning and skill completion.
  {SSP achieves the best performance in single-event settings, while Qwen2.5-Omni-7B performs best in the multi-event setting.}
  \textbf{Best} and \textitbf{second-best} results are highlighted,
  excluding Oracle.
  (b) Progressive completion success rate of object relocation.
  (c) Progressive completion success rate of state transition.
  (d) Progressive completion success rate in the multi-event setting.
  }
  \label{fig:sota}
  \vspace{-0.15in}
\end{figure*}

\noindent\textbf{Quantitative Results of Object Relocation in the Single-Event Setting.}
As shown in~\reffig{fig:sota_overall}, SSP achieves the best non-oracle performance in both planning and completion, with a CSR of 62.51\%, compared with 41.83\% for Qwen3-Omni-30B, the strongest zero-shot planner.
Despite its high planning success, a substantial gap remains between planning and completion, with most execution-stage losses in navigation (see~\reffig{fig:pick_place}).
These results highlight the challenge of across-scene navigation, accounting for unfamiliar structures and object layouts across different scenes.

\noindent\textbf{Quantitative Results of State Transition in the Single-Event Setting.}
SSP again achieves the best non-oracle performance, with a CSR of 39.51\% (see~\reffig{fig:sota_overall}).
The progressive curves in~\reffig{fig:state_transition} also show substantial execution-stage losses in navigation.
These results highlight that state transition in unseen scenes demands both appropriate skill selection and successful navigation to the corresponding sound sources, rather than articulated-object control alone.

\noindent\textbf{Quantitative Results in the Multi-Event Setting.}
Multi-event completion is more challenging, with Oracle achieving a completion success rate of 39.50\% and Qwen2.5-Omni-7B reaching 31.60\% (see~\reffig{fig:sota_overall}).
The progressive curves in~\reffig{fig:2sound} show further reductions as execution proceeds through two navigation skills.
These results highlight two key challenges: generalizing the navigation skill to unseen scenes with diverse structures and layouts, and reliably executing successive navigation and interaction skills to resolve multiple sound events.

\begin{figure}[!t]
    \centering
    \includegraphics[width=0.99\linewidth]{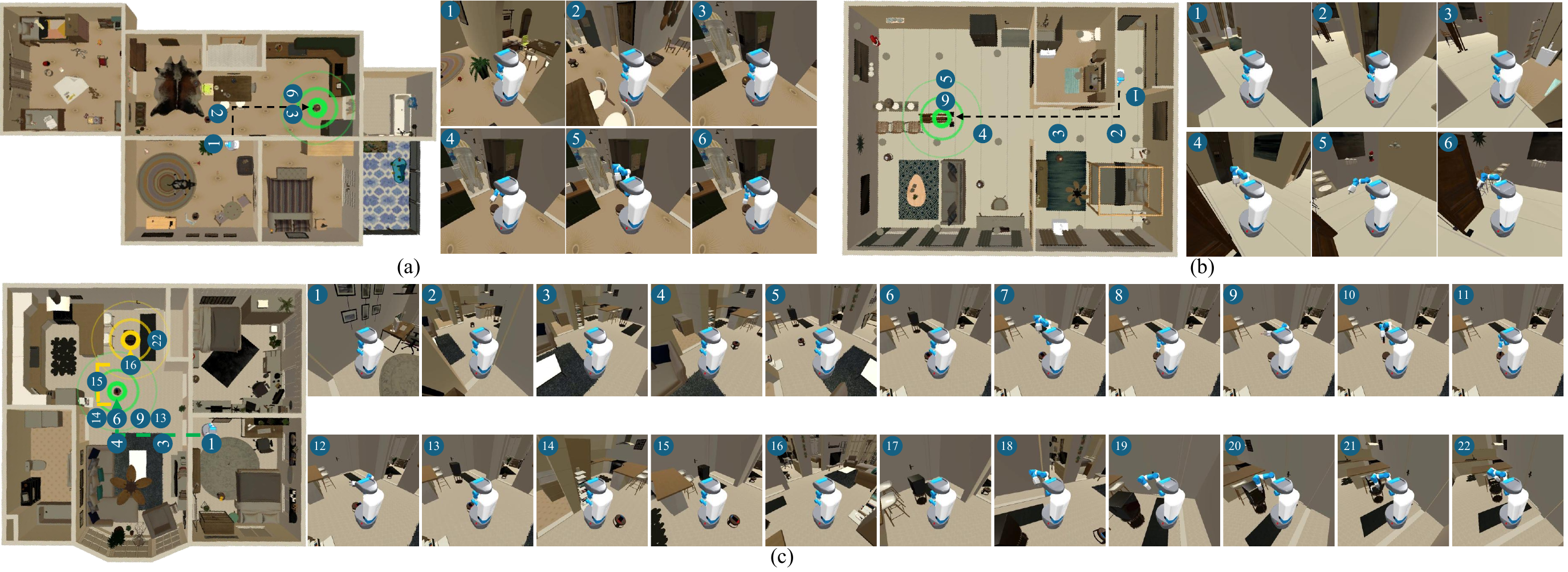}
    \vspace{-1em}
    \caption{
    {Qualitative Visualizations of Task Execution (Zoom in for a better visualization).} 
Execution trajectories for (a) {Object Relocation}, (b) {State Transition}, and (c) {Multi-event Setting}. For each task, the {left panel} maps robot paths and sound sources from a top-down view, while the {right panel} shows third-person keyframes. Multi-event Setting requires sequential navigation and interaction with two distinct sound sources. 
\textit{More visualizations can be seen in the supplementary material.}
}
    \label{fig:visualization}
    \vspace{-.5em}
\end{figure}

\noindent\textbf{Qualitative Result.}
We visualize executions of the robot on the three tasks in~\reffig{fig:visualization}.
In the object relocation, the robot successfully navigates to the target and executes the {Pick} and {Place} skills (see~\reffig{fig:visualization}a).
In the state transition, the robot navigates to the sound-emitting door and performs the {Open Door} skill (see~\reffig{fig:visualization}b).
{The multi-event setting involves two simultaneous sound sources: \textcolor{stmmalarm}{a primary source} and \textcolor{stmmdoorbell}{a secondary source} (see~\reffig{fig:visualization}c).
Following the specified event priority, the robot first interacts with the primary source (switch) despite interference from the secondary signal and then interacts with the secondary source (airfrier). }
More visualizations can be seen in the supplementary material.

\subsection{Ablation Study and Further Discussion}
\noindent\textbf{Comparison with existing navigation policies.}
We compare the proposed navigation policy model with existing ones~\citep{chen_savi_cvpr21,zeng2026savnce} on object relocation (see~\reftab{tab:ablation_nav_audio_a}).
Our model achieves a navigation success rate of 75.28\%, outperforming SAVi (15.68\%) and MAGNet (71.47\%), with performance boosts over the baselines.

\noindent\textbf{Effect of memory components on the navigation policy.}
In~\reftab{tab:ablation_nav_audio_b}, using Explicit Spatial Memory (E.S.M.) or Compressed Spatio-Temporal Memory (C.S.T.M.) alone yields success rates of 36.00\% and 44.08\%, respectively, while combining them achieves 75.28\%.
This performance gain highlights the value of explicitly preserving spatial memory to complement compressed memory.

\noindent\textbf{Effect of input modalities on the open refrigerator policy model.}
For open refrigerator policy training, we evaluate depth inputs from the head and arm viewpoints in~\reftab{tab:ablation_nav_audio_c}.
Head depth alone achieves 94.55\% success, compared with 92.91\% for arm depth alone; combining both increases success to 95.36\%.
Both single-view variants perform well, with the additional arm view providing a modest gain over head depth alone. This performance boost highlights the complementary nature of the two views: the arm view focuses on the interactive regions of the refrigerator, while the head view guides the robot's base movement.

\noindent\textbf{Effect of input modalities on supervised setting-specific planner.}
We conduct ablation studies on the task planning of the state transition.
As shown in~\reftab{tab:ablation_nav_audio_d}, sound alone achieves a success rate of 59.50\%, substantially exceeding the 11.61\% obtained with head RGB alone.
Combining sound and head RGB further improves success to 61.49\%.
These results suggest that sound provides the primary cue for skill-chain prediction, while visual context contributes complementary information.

\begin{table*}[!t]
  \centering
  \caption{
  {Navigation comparison and Ablation Study.}
  (a)~Comparison with existing navigation policies~\citep{chen_savi_cvpr21,zeng2026savnce}.
  (b)~Effect of memory components on the navigation policy.
  E.S.M. and C.S.T.M. denote Explicit Spatial Memory and
  Compressed Spatio-Temporal Memory, respectively.
  (c)~Effect of input modalities on the open refrigerator policy model.
  (d)~Effect of input modalities on the supervised setting-specific planner.
  S.R. denotes success rate.
  }
  \vspace{-1em}
  \label{tab:ablation_all}

  \begingroup
  \small
  \setlength{\tabcolsep}{3pt}
  \renewcommand{\arraystretch}{1.15}

  \captionsetup[subtable]{
    position=top,
    skip=4pt,
    justification=centering,
    font=small
  }

  \resizebox{\linewidth}{!}{%
    \subcaptionbox{\label{tab:ablation_nav_audio_a}}{%
      \begin{tabular}{@{}cc@{}}
        \shline
        Method & S.R. \\
        \hline
        SAVi   & 15.68 \\
        MAGNet & 71.47 \\
        Ours   & \textbf{75.28} \\
        \shline
      \end{tabular}%
    }%
    \hspace{10pt}%
    \subcaptionbox{\label{tab:ablation_nav_audio_b}}{%
      \begin{tabular}{@{}cccc@{}}
        \shline
        Method & E.S.M. & C.S.T.M. & S.R. \\
        \hline
        \Rnum{1}
          & \textcolor{nvidiagreen}{\ding{51}}
          & \textcolor{red}{\ding{55}}
          & {36.00} \\
        \Rnum{2}
          & \textcolor{red}{\ding{55}}
          & \textcolor{nvidiagreen}{\ding{51}}
          & {44.08} \\
        \Rnum{3}
          & \textcolor{nvidiagreen}{\ding{51}}
          & \textcolor{nvidiagreen}{\ding{51}}
          & \textbf{75.28} \\
        \shline
      \end{tabular}%
    }%
    \hspace{10pt}%
    \subcaptionbox{\label{tab:ablation_nav_audio_c}}{%
      \begin{tabular}{@{}cccc@{}}
        \shline
        Method & Head Depth & Arm Depth & S.R. \\
        \hline
        \Rnum{1}
          & \textcolor{red}{\ding{55}}
          & \textcolor{nvidiagreen}{\ding{51}}
          & 92.91 \\
        \Rnum{2}
          & \textcolor{nvidiagreen}{\ding{51}}
          & \textcolor{red}{\ding{55}}
          & 94.55 \\
        \Rnum{3}
          & \textcolor{nvidiagreen}{\ding{51}}
          & \textcolor{nvidiagreen}{\ding{51}}
          & \textbf{95.36} \\
        \shline
      \end{tabular}%
    }%
    \hspace{10pt}%
    \subcaptionbox{\label{tab:ablation_nav_audio_d}}{%
      \begin{tabular}{@{}cccc@{}}
        \shline
        Method & Sound & Head RGB & S.R. \\
        \hline
        \Rnum{1}
          & \textcolor{red}{\ding{55}}
          & \textcolor{nvidiagreen}{\ding{51}}
          & 11.61 \\
        \Rnum{2}
          & \textcolor{nvidiagreen}{\ding{51}}
          & \textcolor{red}{\ding{55}}
          & 59.50 \\
        \Rnum{3}
          & \textcolor{nvidiagreen}{\ding{51}}
          & \textcolor{nvidiagreen}{\ding{51}}
          & \textbf{61.49} \\
        \shline
      \end{tabular}%
    }%
  }
  \endgroup
  \vspace{-.5em}
\end{table*}

\section{Conclusion}
\label{sec:conclusion}
We propose MuSTMM, introducing multiple scenes into sound-triggered mobile manipulation while supporting spatial audio rendering and physical interaction.
The scene-disjoint protocol enables evaluation of single-event and multi-event tasks in unseen scenes.
To improve navigation for subsequent interaction, we augmented the Navigation Policy Model with explicit spatial memory where depth observations are accumulated and represented as a BEV occupancy map. Features are fused with sound context to provide target-location and scene-structure priors.
Navigation comparisons and memory ablations show that explicit spatial memory complements compressed spatio-temporal memory to improve sound-source navigation.
Despite these gains, navigation remains a major execution bottleneck in unseen scenes, particularly for multi-event tasks, motivating further research on reliable multi-scene sound-triggered mobile manipulation.

\section{AI Use Statement}
In this work, we used generative AI tools (\eg, ChatGPT) to edit the research paper to improve readability, specifically for grammar and spelling checks. We have not used generative AI tools for creating scientific figures or any plagiarism. We have carefully reviewed all AI-assisted text revisions to supports their accuracy. We take full responsibility for the final content of this work.

\bibliography{iclr2027_conference}
\bibliographystyle{iclr2027_conference}

\clearpage
\section{Appendix}
\newcommand{\apptocline}[3][0em]{%
  \par\noindent\hspace*{#1}%
  \hyperref[#2]{\makebox[2.2em][l]{\ref*{#2}}#3}
  \nobreak\dotfill\ \makebox[1.8em][r]{\pageref{#2}}\par   
  \vspace{2pt}%
}

\begin{center}{\large\scshape Contents}\end{center}
\vspace{0.4em}\hrule\vspace{0.6em}
{\small
\apptocline{sec:prompt}{Prompt for Task Planning}
\apptocline{sec:ablation}{Additional Ablation Study}
\apptocline{sec:detailed_performance}{Detailed Numerical Results on MuSTMM}
\apptocline{sec:additional_visualization}{Additional Qualitative Result}
\apptocline{sec:episode}{Episode statistic}
}
\vspace{0.4em}\hrule

\subsection{Prompt for Task Planning}
\label{sec:prompt}
We utilize a standardized prompt structure across all task planners for a fair comparison. 
The prompt consists of two components: a {system prompt} and a {user prompt}. 
The system prompt defines the detailed rules and logic for task planning, while the user prompt provides the environmental context, including visual and acoustic observations.

\subsubsection{Prompt for Task Planning in the Single-event Setting.} 
The system prompt for the single-event setting consists of four components. 
The first component defines skill library. 
The second establishes the mapping between sound and the corresponding skill chains. 
The third specifies the planning constraints and the mandatory JSON output format. 
Finally, the fourth component provides a template for the response. 
Complementing this, the user prompt supplies the visual and acoustic observations to start the planning process.

\begin{tcolorbox}[
    breakable,
    title= (\Rnum{1}) Prompt for Task Planning in the Single-event Setting,
    before skip=10pt,      
    lines before break=1, 
    after skip=34pt, 
]

\textbf{System Prompt:} You are an expert Embodied AI Task Planner. Your goal is to generate a strictly ordered list of skills based on the input auditory (sound) and visual (head\_rgb) observations.

\textbf{1. ALLOWED SKILL VOCABULARY:}
You are strictly limited to ONLY the following skills. Do not invent new words or change spellings: [``nav", ``pick", ``place", ``open\_door", ``close\_sink", ``open\_air\_fryer", ``open\_dishwasher", ``open\_microwave", ``open\_refrigerator", ``open\_washing\_machine"].

\textbf{2. LOGIC MAPPING (AUDIO $\rightarrow$ SKILL):} Analyze the input audio sound and classify the source object to determine the skill sequence:
\begin{enumerate}
    \item Category A: Relocation Objects
    \begin{enumerate}
        \item If the sound source is:  Mechanical Alarm, Phone (ringing tone), Furby (toy), Talkie, Switch, Radio, Music Box, Bluetooth speaker, Earbuds, Watch, or Tablet. REQUIRED PLAN: [``nav", ``pick", ``place"]
    \end{enumerate}
    \item Category B: State Transition
    \begin{enumerate}
        \item If the sound source is: Doorbell. REQUIRED PLAN: [``nav", ``open\_door"]
        \item If the sound source is: Water Running. REQUIRED PLAN: [``nav", ``close\_sink"]
        \item If the sound source is: Air Fryer Beeping. REQUIRED PLAN: [``nav", ``open\_air\_fryer"]
        \item If the sound source is: Dishwasher Running. REQUIRED PLAN: [``nav", ``open\_dishwasher"]
        \item If the sound source is: Microwave Beeping. REQUIRED PLAN: [``nav", ``open\_microwave"]
        \item If the sound source is: Refrigerator Alarm. REQUIRED PLAN: [``nav", ``open\_refrigerator"]
        \item If the sound source is: Washing Machine Running. REQUIRED PLAN: [``nav", ``open\_washing\_machine"]
    \end{enumerate}
\end{enumerate}
\textbf{3. CONSTRAINTS and OUTPUT FORMAT:}
\begin{enumerate}
    \item Never return an empty list.
    \item Output MUST be a valid JSON object containing a single key ``plan".
    \item Do not include markdown formatting (like \texttt{```} json), just the raw JSON string.
\end{enumerate}
\textbf{4. OUTPUT FORMAT: } \{``plan": [``skill\_1", ``skill\_2", ...]\}
\newline

\textbf{User Prompt:} 
\begin{enumerate}
    \item \textsc{Image Input}
    \item \textsc{Audio Input}
    \item ``Input observation provided. Based on the audio class and visual context, output the execution plan JSON."
\end{enumerate}

\end{tcolorbox}

\subsubsection{Prompt for Task Planning in the Multi-event Setting.} 
For the multi-event setting, the system prompt is composed of five segments to handle concurrent sound sources.
The initial segment lists the permissible skill vocabulary.
The second details the correspondence between sound categories and their required skill sequences.
Crucially, the third segment enforces {interaction priority}. We explicitly identify the primary sound source, requiring the planner to deduce the secondary source from the mixed audio and generate a sequential execution plan for both.
The fourth and fifth segments impose strict output constraints and provide a valid JSON template to ensure parsing reliability.
Finally, the user prompt provides the model with multimodal context, including visual frames and sound data.
\begin{tcolorbox}[
    breakable,
    title= (\Rnum{2}) Prompt for Task Planning in the Multi-event Setting,
    before skip=10pt,      
    lines before break=1, 
    after skip=34pt, 
]

\textbf{System Prompt:} You are an expert Embodied AI Task Planner. Your goal is to generate a strictly ordered list of skills based on the input auditory and visual observations. There are exactly two sound sources detected.

\textbf{1. ALLOWED SKILL VOCABULARY (STRICT):}
You are strictly limited to ONLY the following skills: [``nav", ``pick", ``place", ``open\_door", ``close\_sink", ``open\_air\_fryer", ``open\_dishwasher", ``open\_microwave", ``open\_refrigerator", ``open\_washing\_machine"]. Do not invent new words or change spellings.

\textbf{2. LOGIC MAPPING (AUDIO $\rightarrow$ SKILL):} Analyze the input audio sound to determine the action sequence:
\begin{enumerate}
    \item Category A: Relocation Objects
    \begin{enumerate}
        \item If the sound source is:  Mechanical Alarm, Phone (ringing tone), Furby (toy), Talkie, Switch, Radio, Music Box, Bluetooth speaker, Earbuds, Watch, or Tablet. REQUIRED PLAN: [``nav", ``pick", ``place"]
    \end{enumerate}
    \item Category B: State Transition
    \begin{enumerate}
        \item If the sound source is: Doorbell. REQUIRED PLAN: [``nav", ``open\_door"]
        \item If the sound source is: Water Running. REQUIRED PLAN: [``nav", ``close\_sink"]
        \item If the sound source is: Air Fryer Beeping. REQUIRED PLAN: [``nav", ``open\_air\_fryer"]
        \item If the sound source is: Dishwasher Running. REQUIRED PLAN: [``nav", ``open\_dishwasher"]
        \item If the sound source is: Microwave Beeping. REQUIRED PLAN: [``nav", ``open\_microwave"]
        \item If the sound source is: Refrigerator Alarm. REQUIRED PLAN: [``nav", ``open\_refrigerator"]
        \item If the sound source is: Washing Machine Running. REQUIRED PLAN: [``nav", ``open\_washing\_machine"]
    \end{enumerate}
\end{enumerate}
\textbf{3. OBJECT MAPPING INSTRUCTION (CRITICAL):}
You are provided with an event-priority specification: One of the sound sources has a high priority. You have to interact with ``\texttt{\{obj\_1\}}" first. Your task is to identify the other sound source from the audio and generate the plan for the whole process. You MUST map the objects exactly as follows:
\begin{enumerate}
    \item Key ``first\_sound": Generate the plan for ``\texttt{\{obj\_1\}}".
    \item Key ``second\_sound": Identify the SECOND sound source from the audio (the one that is NOT ``\texttt{\{obj\_1\}}") and generate its plan.
    \item Do NOT output plans for any other potential objects unless they are detected as the second sound.
\end{enumerate}
\textbf{4. CONSTRAINTS and OUTPUT FORMAT:}
\begin{enumerate}
    \item Output MUST be a valid JSON object containing a single root key ``plan".
    \item The keys inside ``plan" MUST be exactly {``}first\_sound" and ``second\_sound".
    \item Do not include markdown formatting (like \texttt{```} json), just the raw JSON string.
\end{enumerate}
\textbf{5. Expected JSON Structure:} \{
        ``plan": \{
            
            ``first\_sound": [``skill\_1", ``skill\_2", ...],  \# $\leftarrow$ Actions for the object assigned to 'first\_sound'

            ``second\_sound": [``skill\_1", ``skill\_2", ...]  \# $\leftarrow$ Actions for the object assigned to 'second\_sound'
        
        \}
    \}
\newline

\textbf{User Prompt:} 
\begin{enumerate}
    \item \texttt{Image Input}
    \item \texttt{Audio Input}
    \item ``Input observation provided. Based on the audio class and visual context, output the execution plan JSON."
\end{enumerate}
\end{tcolorbox}
\subsection{Additional Ablation Study}
\label{sec:ablation}
We evaluate the impact of different input modalities for the Setting-Specific Planner (SSP) in the multi-event setting (see~\reftab{tab:ablation_task_planning}).
Unlike in the single-event state-transition setting, the sound-only variant performs best in the multi-event setting, achieving 75.40\%, compared with 71.70\% when head RGB is also included.
\begin{table}[h]
  \centering
  \caption{Ablation study on input modalities for Setting-Specific Planner (SSP) in the multi-event setting.}
    \resizebox{0.4\linewidth}{!}{
    \begin{tabular}{cccc}
    \shline
    Method & Sound & Head RGB & S.R. \\
    \hline
    I     & \textcolor{red}{\ding{55}}     & \textcolor{nvidiagreen}{\ding{51}}     & 51.00  \\
    II   & \textcolor{nvidiagreen}{\ding{51}}     & \textcolor{nvidiagreen}{\ding{51}}     & {71.70}  \\
    III    & \textcolor{nvidiagreen}{\ding{51}}     & \textcolor{red}{\ding{55}}     & \pmb{75.40}  \\
    \shline
    \end{tabular}}%
  \label{tab:ablation_task_planning}%
\end{table}%

\subsection{Detailed Numerical Results on MuSTMM}
\label{sec:detailed_performance}
We provide detailed numerical results on MuSTMM, including individual skill success rates and conditional skill completion rates (see~\reftab{tab:main_results}).
The conditional success rate of skill $S_i$ measures the probability of successfully completing $S_i$, given a correct task plan and successful completion of all preceding skills.
The Progressive Skill Completion Success Rate (PSCSR) at stage $k$ measures the fraction of episodes that successfully complete task planning and the first $k$ skills.
The Completion Success Rate (CSR) measures the fraction of episodes that successfully complete the entire task.
PSCSR and CSR are defined as
\begin{equation}
\begin{aligned}
\mathrm{PSCSR}_k
&= SR_{\mathrm{plan}}
\prod_{i=1}^{k}
P(S_i \mid S_{1:i-1}, \mathcal{C}), \\
\mathrm{CSR}
&= \mathrm{PSCSR}_K
= SR_{\mathrm{plan}}
\prod_{i=1}^{K}
P(S_i \mid S_{1:i-1}, \mathcal{C}),
\end{aligned}
\label{eq:pscsr}
\end{equation}
where $SR_{\mathrm{plan}}$ denotes the planning success rate, $\mathcal{C}$ indicates a correct task plan, $S_i$ denotes the successful completion of the $i$-th skill, and $K$ is the total number of skills in the task chain.
Thus, $\mathrm{PSCSR}_0=SR_{\mathrm{plan}}$, and CSR corresponds to the final PSCSR after all skills have been executed.
In Table~3, skill columns report conditional success rates, while the progressive completion curves in Figure~4 show PSCSR.
In the multi-event setting, $--$ denotes no surviving episodes for cascade evaluation.

\begin{table*}[!t]
  \centering
  \small
  \caption{{Numerical Results on the MuSTMM.}
  We report success rates ($\%$) for task planning ($SR_{\text{plan}}$), individual skill evaluation (Individual), conditional skill completion (skill columns), and full-task completion (CSR). The higher the better.
  {(a)} {Evaluation of object relocation}. 
  {(b)} {Evaluation of state transition}. 
  {(c)} {Evaluation of multi-event setting}. Success rates are reported in the format ``First Sound Source / Second Sound Source''.
  }
  \vspace{-1em}
  \label{tab:main_results}
    \begin{subtable}[t]{0.47\linewidth}
    \centering
    \caption{}
    \resizebox{\linewidth}{!}{%
    \begin{tabular}{cccccc}
    \shline
    Method & $SR_{\text{plan}}$ & Navigate & Pick & Place & CSR \\
    \hline
    Individual & N/A & 75.28 & 89.48 & 93.82 & N/A \\
    Oracle & 100.00 & 75.28 & 88.87 & 99.64 & 66.67 \\
    \hline
    \hline
    Qwen2.5-Omni-3B & 0.36 & 100.00 & 83.33 & 100.00 & 0.30 \\
    Qwen2.5-Omni-7B & 5.11 & 56.98 & 89.80 & 100.00 & 2.61 \\
    MiniCPM-o-4.5 & 11.76 & 69.70 & 92.03 & 99.21 & 7.49 \\
    Qwen3-Omni-30B & 63.76 & 74.74 & 88.03 & 99.72 & 41.83 \\
    SSP & 94.00 & 75.22 & 88.66 & 99.72 & \textbf{62.51} \\
    \shline
    \end{tabular}%
    }
    \label{tab:sota_stow}
    \end{subtable}%
    \hfill
    \begin{subtable}[t]{0.47\linewidth}
    \centering
    \caption{}
    \resizebox{\linewidth}{!}{%
    \begin{tabular}{ccccc}
    \shline
    Method & $SR_{\text{plan}}$ & Navigate & Interaction & CSR \\
    \hline
    Individual & N/A & 68.28 & 91.73 & N/A \\
    Oracle & 100.00 & 68.28 & 92.43 & 63.12 \\
    \hline
    \hline
    Qwen2.5-Omni-3B & 18.44 & 69.18 & 98.44 & 12.56 \\
    Qwen2.5-Omni-7B & 32.31 & 72.32 & 95.58 & 22.34 \\
    MiniCPM-o-4.5 & 30.33 & 70.51 & 94.80 & 20.27 \\
    Qwen3-Omni-30B & 38.35 & 71.81 & 95.96 & 26.43 \\
    SSP & 61.49 & 70.07 & 91.70 & \textbf{39.51} \\
    \shline
    \end{tabular}%
    }
    \label{tab:sota_interact}
    \end{subtable}%
    
    \vspace{1em}
    \begin{subtable}[t]{0.90\linewidth}
    \centering
    \caption{}
    \resizebox{\linewidth}{!}{%
    \begin{tabular}{ccccccc}
    \shline
    Method & $SR_{\text{plan}}$ & Navigate & Pick & Place & Interaction & CSR \\
    \hline
    Individual & N/A & 49.20 / 65.10 & 92.32 / 93.70 & 97.55 / 98.35 & 95.16 / 68.75 & N/A \\
    Oracle & 100.00 & 49.20 / 84.38 & 97.91 / 97.67 & 100.00 / 100.00 & 84.62 / 94.74 & 48.00 / 39.50 \\
    \hline
    \hline
    Qwen2.5-Omni-3B & 65.90 & 52.50 / 85.21 & 97.95 / 98.24 & 100.00 / 100.00 & 80.00 / 100.00 & 33.80 / 28.30 \\
    Qwen2.5-Omni-7B & 76.10 & 50.46 / 85.64 & 97.92 / 98.11 & 100.00 / 100.00 & -- / 100.00 & \textbf{37.60} / \textbf{31.60} \\
    MiniCPM-o-4.5 & 33.80 & 51.48 / 90.06 & 98.28 / 95.89 & 100.00 / 100.00 & -- / 87.50 & 17.10 / 14.70 \\
    Qwen3-Omni-30B & 67.20 & 48.66 / 85.67 & 98.14 / 96.93 & 100.00 / 100.00 & 100.00 / 92.86 & 32.10 / 26.60 \\
    SSP & 75.40 & 50.53 / 84.05 & 97.55 / 98.39 & 100.00 / 100.00 & 84.62 / -- & 37.00 / 30.60 \\
    \shline
    \end{tabular}%
    }
    \label{tab:sota_two_sound}
    \end{subtable}%

\end{table*}

\subsection{Additional Qualitative Result}
\label{sec:additional_visualization}
\begin{figure*}[!t]
    \centering
    \includegraphics[width=0.99\linewidth]{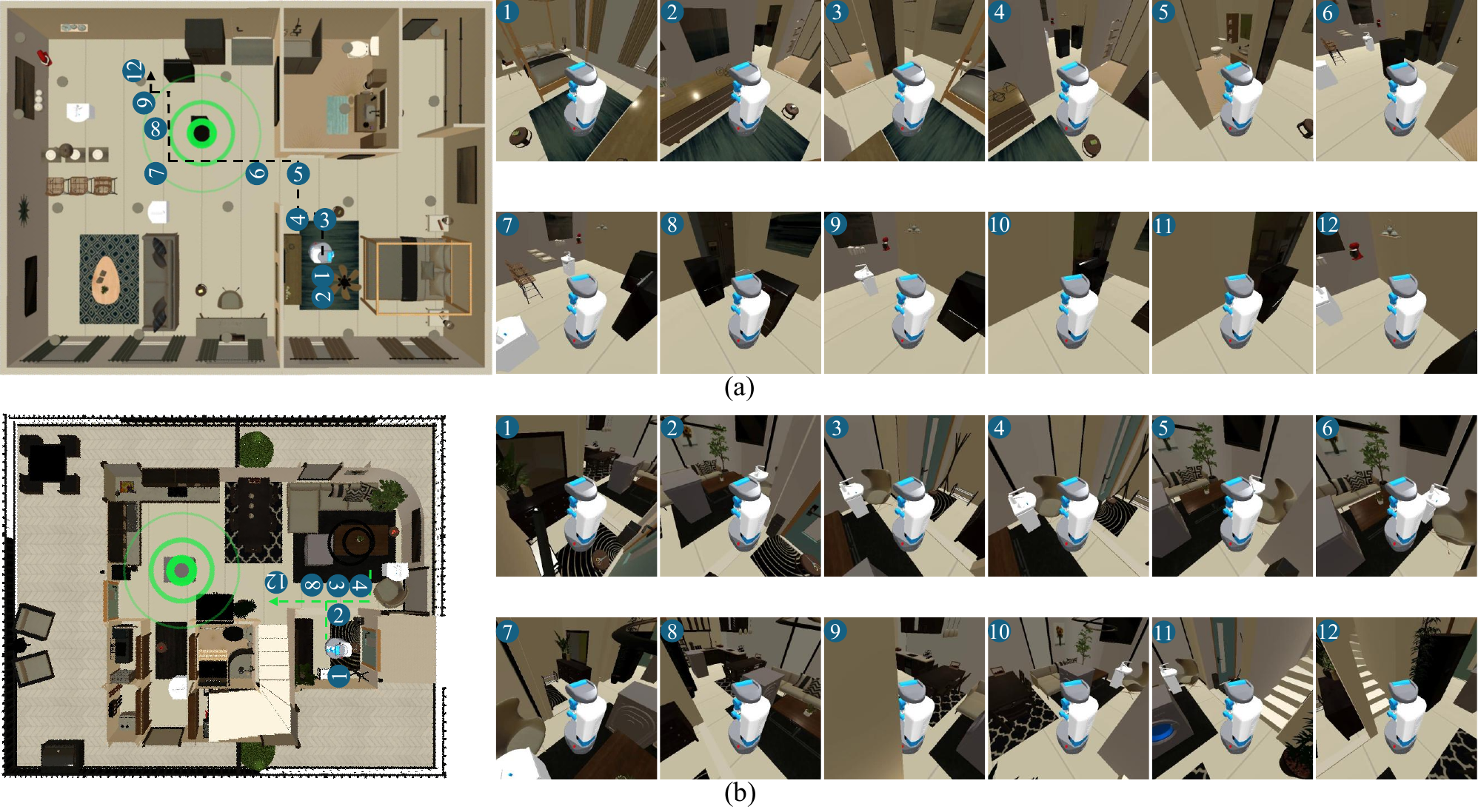}
    \caption{
    Failure Case Visualizations of Navigation Skill Execution (Zoom in for a better visualization).
    For each case, the {left panel} maps robot paths and sound sources from a top-down view, while the {right panel} shows third-person keyframes.
    }
    \label{fig:qa_supp}
\end{figure*}

We present failure cases for the navigation skill, identifying future directions for MuSTMM.
In \reffig{fig:qa_supp}a, the robot stops near a silent, visually distracting refrigerator instead of the target one. This case highlights the challenges from visual distractors, specifically the presence of intra-class ones.
In \reffig{fig:qa_supp}b, the robot is initialized in a highly confined space. Notably, it successfully navigates out of this initial confinement and an additional trap, showing robust spatial recovery capabilities of the proposed navigation policy. However, due to the complex architectural structure of the scene and misleading acoustic echoes, the robot is temporarily diverted in the opposite direction (wall). Although it eventually corrects its trajectory and resumes moving toward the correct target, the episode terminates upon reaching the maximum step limit. These challenging cases show that while the proposed navigation skill can effectively escape confined spaces, handling extreme visual ambiguity and intricate acoustic environments remains a valuable direction for future research.
\textit{More visualizations can be seen in the attached videos.}

\subsection{Episode statistic}
\label{sec:episode}
\begin{figure*}[!t]
    \centering
    \includegraphics[width=0.4\linewidth]{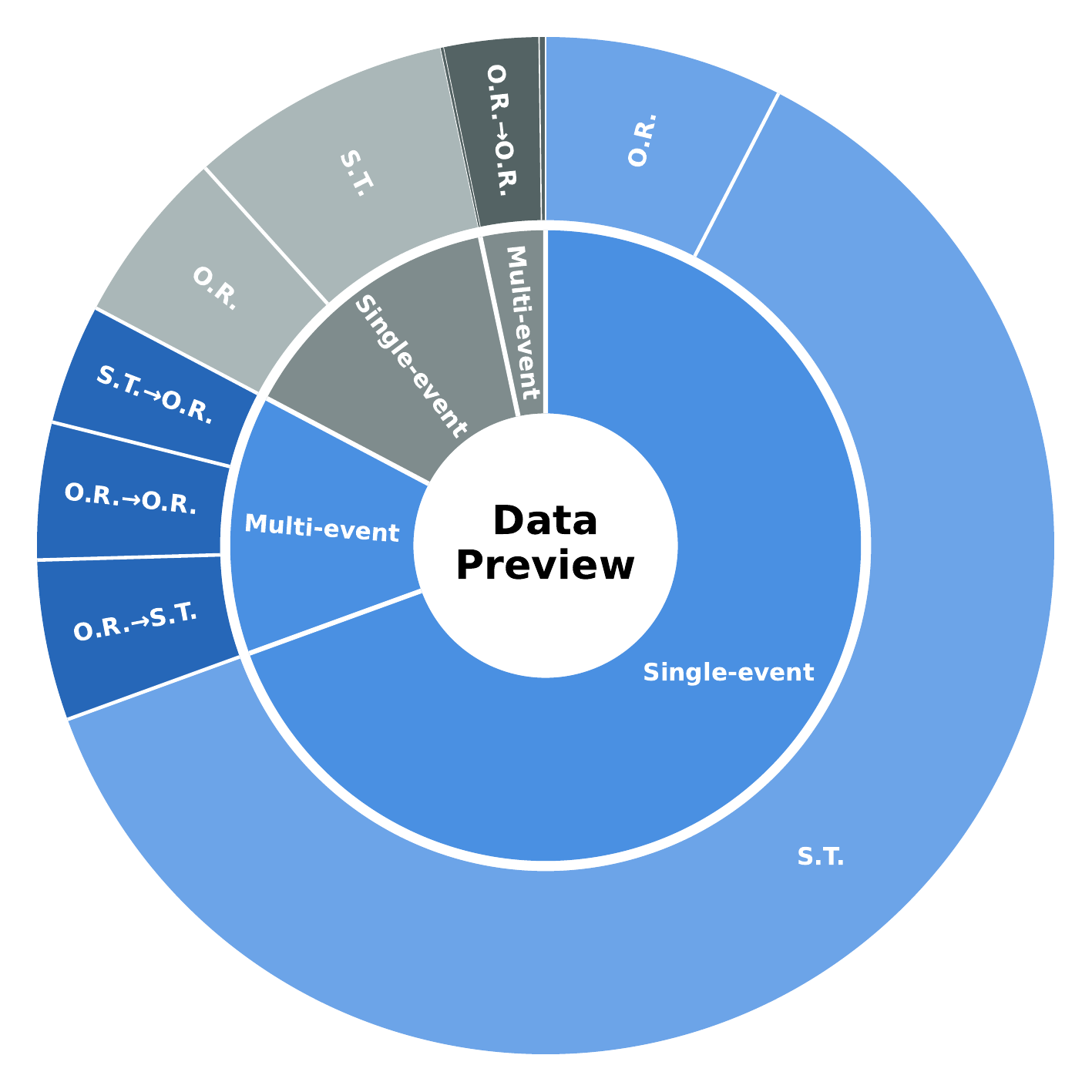}
    \caption{
    \textcolor{TrainColor}{Train} and \textcolor{TestColor}{test} splits. (\textcolor{TrainColor}{24.91k} and \textcolor{TestColor}{5.20k} episodes)
        O.R. denotes Object Relocation, and S.T. denotes State Transition.
        Interaction type of the first sound source $\rightarrow$ the second one.
    }
    \label{fig:episode_data}
\end{figure*}
Episode statistics are shown in~\reffig{fig:episode_data}. We exclude episodes involving two articulated sound sources during data generation, as structural variations make it difficult to reliably keep safe interaction regions for articulated objects.

\end{document}

%% file: math_commands.tex
\usepackage{amsmath,amsfonts,bm}

\def\eqref#1{equation~\ref{#1}}

\def\1{\bm{1}}

\DeclareMathAlphabet{\mathsfit}{\encodingdefault}{\sfdefault}{m}{sl}
\SetMathAlphabet{\mathsfit}{bold}{\encodingdefault}{\sfdefault}{bx}{n}

